\documentclass[10pt,twocolumn,letterpaper]{article}

\usepackage{iccv}              

\newcommand{\tablestyle}[2]{\setlength{\tabcolsep}{#1}\renewcommand{\arraystretch}{#2}\centering\footnotesize}
\def\module{MagicID}
\def\modulespace{MagicID }

\definecolor{iccvblue}{rgb}{0.21,0.49,0.74}
\usepackage[pagebackref,breaklinks,colorlinks,allcolors=iccvblue]{hyperref}

\usepackage{amsmath}
\usepackage{amssymb}
\usepackage{mathtools}
\usepackage{amsthm}
\usepackage{graphicx}

\usepackage{caption}
\usepackage{subcaption}

\usepackage{hyperref}
\usepackage{url}
\usepackage{booktabs} 
\usepackage{wrapfig}

\title{\module: Hybrid Preference Optimization for ID-Consistent and Dynamic-Preserved Video Customization}

\author{%
Hengjia Li$^{1,*}$, Lifan Jiang$^{1,*}$, Xi Xiao$^{2,*}$, Tianyang Wang$^{2}$, Hongwei Yi$^{3}$, Boxi Wu$^{1}$, Deng Cai$^{1}$ 
 \\\\
 $^1$Zhejiang University\qquad$^2$University of Alabama at Birmingham\qquad
 $^3$Hedra AI
 \\ 
 {\tt\small lihengjia98@gmail.com}\\ 
 Project page: \url{ https://echopluto.github.io/MagicID-project/}
}

\usepackage[misc]{ifsym}
\newcommand\blfootnote[1]{
    \begingroup
    \renewcommand\thefootnote{}\footnote{#1}
    \addtocounter{footnote}{-1}
    \endgroup
}
\usepackage[hang]{footmisc}

\begin{document}

\twocolumn[{
\renewcommand\twocolumn[1][]{#1}
\maketitle
\begin{center}
    \centering
    \vspace*{-3mm}
    \includegraphics[width=\textwidth]{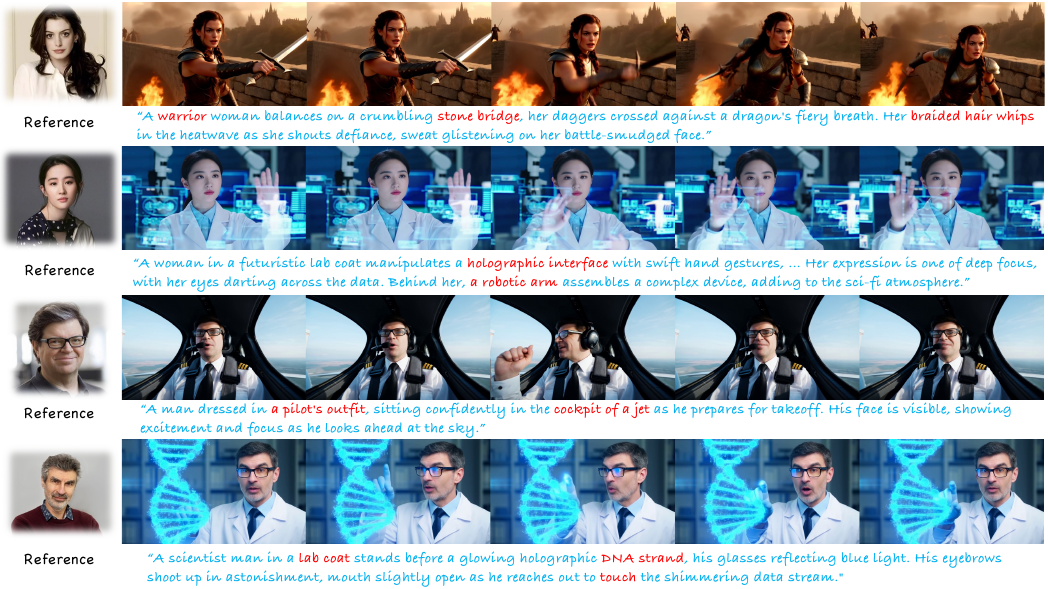}

        \captionof{figure}{
        \textbf{Results of \module.} Given a few reference images, our method is capable of generating highly realistic and personalized videos that maintain consistent identity features while exhibiting natural and visually appealing motion dynamics. 
      }
\label{fig:teaser}
\end{center}
}]

{
    \blfootnote{
        * Equal Contribution        }
}

\begin{abstract}

Video identity customization seeks to produce high-fidelity videos that maintain consistent identity and exhibit significant dynamics based on users' reference images. However, existing approaches face two key challenges: identity degradation over extended video length and reduced dynamics during training, primarily due to their reliance on traditional self-reconstruction training with static images. To address these issues, we introduce \textbf{\module}, a novel framework designed to directly promote the generation of identity-consistent and dynamically rich videos tailored to user preferences. Specifically, we propose constructing pairwise preference video data with explicit identity and dynamic rewards for preference learning, instead of sticking to the traditional self-reconstruction. To address the constraints of customized preference data, we introduce a hybrid sampling strategy. This approach first prioritizes identity preservation by leveraging static videos derived from reference images, then enhances dynamic motion quality in the generated videos using a Frontier-based sampling method. By utilizing these hybrid preference pairs, we optimize the model to align with the reward differences between pairs of customized preferences. Extensive experiments show that \modulespace successfully achieves consistent identity and natural dynamics, surpassing existing methods across various metrics.
\end{abstract}    
\section{Introduction}
\label{sec:intro}

Identity-specific content generation~\citep{gal2022image, ruiz2023dreambooth, ye2023ip, li2024photomaker, wang2024instantid, guo2024pulid, ma2024magic, he2024id, wu2024customcrafter, wei2024dreamvideo, yuan2024identity, li2024personalvideo} has long been a central focus in the field of computer vision. Recent breakthroughs with the advent of diffusion models, have elevated personalized content creation to a prominent position especially in identity-consistency image generation. However, replicating this level of accuracy in video generation~\citep{guo2023animatediff,wang2023modelscope,chen2024videocrafter2,videoworldsimulators2024, kong2024hunyuanvideo} continues to pose significant challenges. It seeks to tailor a diverse array of captivating videos utilizing a limited set of user's images, enabling the production of personalized video that showcases individuals in various actions and settings, all while ensuring exceptional identity (ID) fidelity. This innovation holds significant promise for applications within the film and television sectors.

\begin{figure}[t]
    \centering
    \begin{subfigure}{0.48\linewidth}
        \centering
        \includegraphics[width=\linewidth]{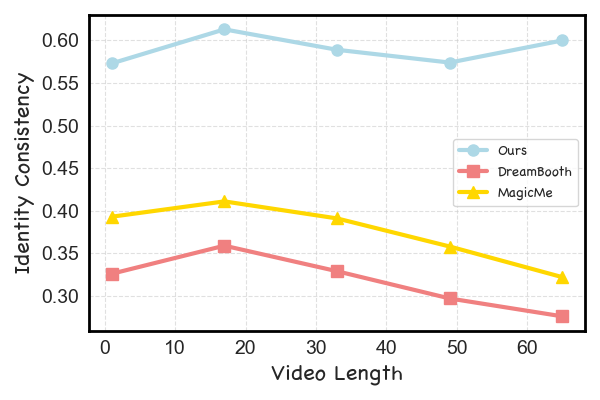}
        \caption{ID Consistency for longer video.}
        \label{fig:identity}
    \end{subfigure}
    \hfill
    \begin{subfigure}{0.48\linewidth}
        \centering
        \includegraphics[width=\linewidth]{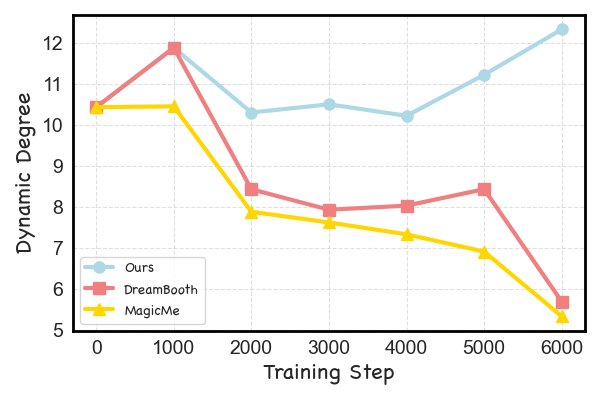}
        \caption{Dynamic Degree for more steps.}
        \label{fig:dynamic}
    \end{subfigure}
    \caption{\textbf{Analysis of identity degratation and dynamic reduction.} (a) We compute the mean identity similarity with the reference images for generated videos of different lengths. As shown, traditional approaches suffer from diminished identity consistency as video length increases. In contrast, our method maintains strong identity robustness throughout prolonged video generations. (b) We calculate the dynamic degree for different training steps. As the customization progresses, traditional methods experience a gradual loss of motion dynamic during customization, whereas our method preserves original video dynamics across the entire training.}
    \label{fig:identity_dynamic}
\end{figure}

Compared to image customization, the challenge of video customization lies in utilizing a limited number of static images rather than videos as references. Following the conventional approach of image customization, existing video customization methods~\cite{ma2024magic, wu2024customcrafter, wei2024dreamvideo} naively employ self-reconstruction of reference images to maintain the identity. However, we observe that this process introduces two significant issues, \ie, identity degradation for more frames and dynamic reduction during the training. (1) \textbf{The longer the video length, the worse the identity consistency becomes.} 
Since the reference images are essentially single-frame representations, the inherent disparity in temporal resolution with video sequences (comprising multiple frames) creates a fundamental domain shift. Conventional self-reconstruction training paradigms fail to bridge this gap, resulting in inconsistent performance between training and inference phases. As illustrated in \cref{fig:identity}, models trained with conventional methods exhibit poorer ID consistency when sampling videos with a higher number of frames in the inference phase. (2) \textbf{The longer the training duration, the less the video dynamics become.} Since traditional methods merely reconstruct static images during training, it typically leads to a severe distribution shift, which drives the model to generate videos with less dynamic degree. As shown in \cref{fig:dynamic}, traditional methods demonstrate increasingly inferior dynamics over the course of the customization.

To solve these problems, we propose \module, a novel framework to maintain consistent identity with the reference images and preserve the natural motion dynamics. Inspired by the Direct Preference Optimization for language~\cite{kim2024sdpo,xu2024dpo} and image generation~\cite{liu2024videodpo,jiang2025huvidpo}, we introduce a hybrid customized preference optimization to directly encourage the model to generate ID-consistent and dynamic-preserved videos. Instead of previous self-reconstruction approach on reference images, we propose to construct pairwise hybrid-preference video data with explicit customized rewards for the training. 

However, it is challenging to obtain preference video data that simultaneously preserves consistent identity and exhibits good dynamics for the preference learning.
To overcome this limitation, we introduce a hybrid sampling strategy to respectively construct identity-prefered and dynamic-prefered training pairs for two stages. (1) Intuitively, the simplistic utilization of generated sample pairs will constrain the model's ability to assimilate the identity in the invisible references. Therefore, in the first identity-prefered stage, we incorporate the static videos inflated from the reference images with a selection of videos generated by the preliminary model into the base dataset repository. Since the objective of this stage is to maximally incentivize the model to preserve the identity consistently with the reference, thereby we prioritize the disparity in identity reward while permitting a degree of tolerance for dynamic reward. (2) In the subsequent dynamic-prefered stage, we aim to construct Pareto-optimal pairs of dynamics and identity to enhance dynamics while maintaining identity learning. Thus, we incorporate the samples generated by the fine-tuned model into the base dataset repository and construct the preference pairs from the total base dataset repository. Then we introduce a Frontier-based sampling method to select training pairs from the upper and lower Pareto frontiers according to the dynamic and identity rewards, ensuring the learning of both customized preferences.

We evaluate our methodology using a variety of general metrics and compare it with prior competitive methods for identity-preserved video generation. Comprehensive experimental and visual analyses reveal that our technique effectively produces high-quality videos featuring dynamic content and robust facial consistency, as depicted in \cref{fig:teaser}. Our contributions are summarized as follows:
\begin{itemize}
\item We introduce \module, a novel preference optimization to effectively solve the degradation of identity and dynamics. To the best of our knowledge, this is the first work to effectively achieve customized video generation with direct preference optimization.
\item We propose a hybrid sampling strategy to construct high-quality preference video pairs, which addresses the issue of lacking the preference data with consistent identity and considerable dynamics.
\item We demonstrate the effectiveness of \modulespace with favorable video customization quality against state-of-the-art models within a variety of general metrics.
\end{itemize}

\begin{figure*}[!t]
    \centering
    \includegraphics[width=\linewidth]{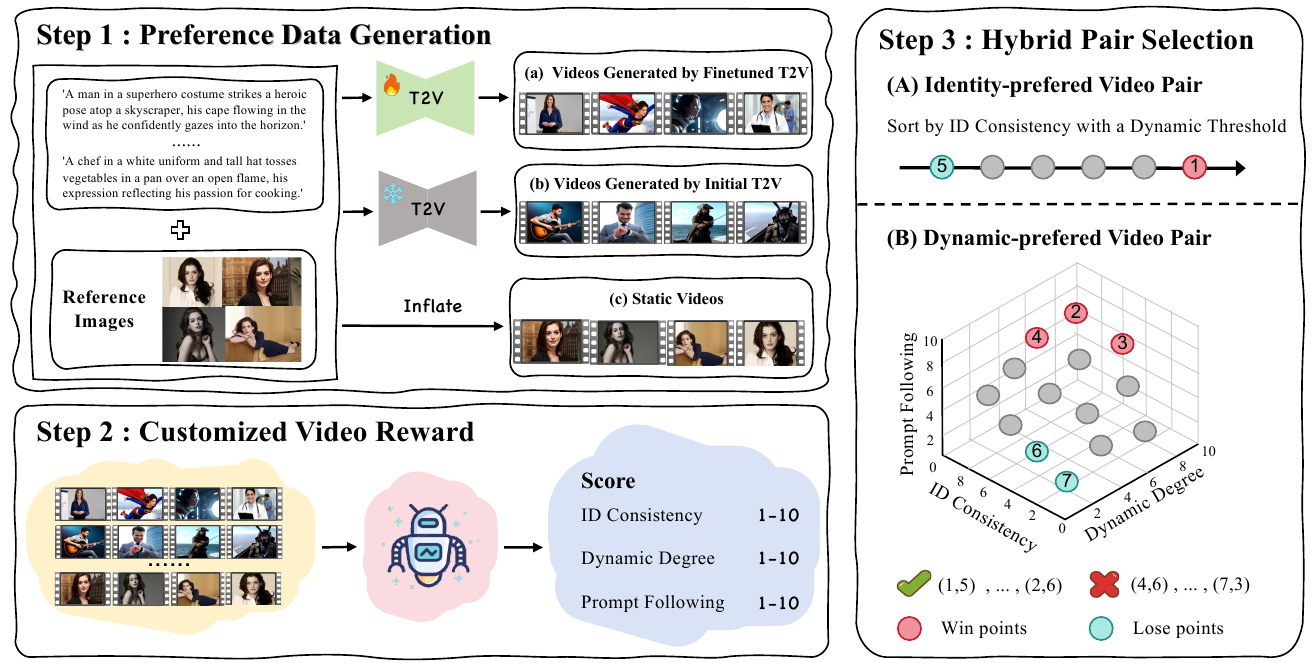}
    \caption{\textbf{Overview of pairwise preference video data construction.} In Step 1, we construct a preference video repository using videos generated by fine-tuned and Initial T2V models, along with static videos derived from reference images. In Step 2, we evaluate each video sequentially based on ID consistency using ID Encoder~\cite{deng2019arcface}, dynamic degree using optical flow~\cite{huang2024vbench}, and prompt following using VLM\cite{bai2023qwenvlversatilevisionlanguagemodel}. In Step 3, we perform Hybrid Pair Selection, first selecting pairs based on ID consistency differences with a pre-defined dynamic threshold to address identity inconsistency, then selecting pairs based on both dynamic and identity to mitigate the dynamic reduction.}
    \label{fig:method}
\end{figure*}

\section{Related Works}
\label{sec:formatting}

\noindent
\textbf{Text-to-Image Customization.}
In the realm of Text-to-Image (T2I) generation, numerous methodologies~\citep{gal2022image, li2024photomaker, gal2024lcm, valevski2023face0, xiao2024fastcomposer, ma2024subject, peng2024portraitbooth, li2023few, li2024unihda} have surfaced for identity (ID) customization. As a foundational work, Textual Inversion~\citep{gal2022image} encodes the user-provided identity into a specific token embedding within a frozen T2I model. To enhance ID fidelity, DreamBooth~\citep{ruiz2023dreambooth} refines the original model, with efficient fine-tuning techniques like LoRA~\citep{hu2021lora} also being applicable. Conversely, encoder-based methods strive to directly infuse ID into the generation process. PhotoMaker~\citep{li2024photomaker} proposes to enhance the ID embedding based on large-scale datasets comprising diverse images of each ID.  PuLID~\citep{guo2024pulid} proposes optimizing an ID loss between the generated and reference images in a more precise configuration.

\noindent
\textbf{Text-to-Video Customization.}
Text-to-Video (T2V) customization introduces additional complexities compared to Text-to-Image (T2I) customization, primarily due to the temporal motion dynamics inherent in videos. Presently, only a limited number of studies~\citep{ma2024magic, he2024id, wu2024customcrafter, wei2024dreamvideo, yuan2024identity, li2024personalvideo} have conducted preliminary explorations in this domain. MagicMe~\citep{ma2024magic} employs an identity (ID) module based on an extended Textual Inversion approach. Nonetheless, training under self-reconstruction on the reference images diverges the videos in the inference, resulting in suboptimal ID fidelity and model degradation. 
ID-Animator~\citep{he2024id} and ConsisID~\cite{yuan2024identity} suggests encoding ID-relevant information with a face adapter, which demands thousands of high-quality human videos for fine-tuning, thereby imposing substantial costs related to dataset construction and model training.

\noindent
\textbf{Direct Preference Optimization.}
Direct Preference Optimization (DPO)~\cite{rafailov2023direct} has emerged as a promising alternative to traditional Reinforcement Learning from Human Feedback (RLHF)~\cite{ziegler1909fine}, as it eliminates the need for training a separate reward model. Initially, DPO was widely adopted in LLMs~\cite{stiennon2020learning,kim2024sdpo,xu2024dpo} to better align model outputs with human preferences. More recently, its application has extended to stable diffusion and related generative tasks. In particular, some studies~\cite{lee2025calibrated,wallace2024diffusion} successfully applied DPO to text-to-image generation, significantly improving the aesthetic quality of 2D-generated images. Furthermore, DreamDPO~\cite{zhou2025dreamdpo} extended DPO to text-to-3D tasks, enhancing both the quality and controllability of generated 3D objects. Similarly, some studies~\cite{liu2024videodpo,jiang2025huvidpo} have introduced DPO to text-to-video generation, leading to notable improvements in video generation performance. Despite DPO's success, its application to personalized video generation remains challenging due to the task's unique nature. There is an urgent need to apply DPO to address identity inconsistency and dynamic loss.

\section{Method}

\subsection{Preliminary}

Direct Preference Optimization implicitly optimizes a policy based on human preference data while constraining deviation from a given reference policy \(\pi_{\text{ref}}\) via a Kullback-Leibler divergence~\cite{kullback1951information}. Although DPO does not explicitly maximize a predefined reward function \(r(x,y)\), its intrinsic optimization objective can still be represented as:
\begin{equation}
\label{max}
\max_\pi \mathbb{E}_{x \in X,\, y \sim \pi(\cdot|x)} \left[ r(x,y) \right] - \beta \, \mathbb{D}_{\text{KL}}\left[ \pi(y|x) \,\|\, \pi_{\text{ref}}(y|x) \right],
\end{equation}
where \(r(x,y)\) is the implicit reward derived from preference data and \(\beta\) controls the divergence from \(\pi_{\text{ref}}\). In practical applications, DPO explicitly learns the relative preference between a preferred output \(y_w\) and a less-preferred output \(y_l\) by employing a cross-entropy loss function derived from the Bradley-Terry model:

\begin{equation} \label{www}
\begin{aligned}
\mathcal{L}_{DPO} = -\mathbb{E}_{(x, y_w, y_l) \sim D} \Bigg[ 
& \ln \sigma\!\Bigg( 
\beta \log \frac{\pi(y_w|x)}{\pi_{\text{ref}}(y_w|x)}  \\
& \quad - \beta \log \frac{\pi(y_l|x)}{\pi_{\text{ref}}(y_l|x)}
\Bigg) \Bigg].
\end{aligned}
\end{equation}

\subsection{Preference Data Generation}
\label{pref}
In this work, we aim to directly optimize the model to generate identity-consistent and dynamic-preserved videos with the preference video data. However, the key difference between our customization task and other generation tasks lies in the fact that we can only obtain prior information from a few reference images provided by the user. As a result, we lack high-quality training data with good identity consistency, dynamic degree, and prompt following, which do not meet the traditional data requirements for preference optimization, posing a significant challenge to the fine-tuning process. To overcome this issue, we propose to construct the base dataset repository through three different approaches.

First, we sample several prompts from Large Language Model and feed them into the video generator $\pi_{\text{ref}}$ with the initial LoRA fine-tuning to generate video data $V_t$. On the other hand, we extend the preference data with videos $V_s$ generated by the original model without LoRA finetuning. These videos represent the original distribution of the T2V model, which remains unaffected by the customization process. However, the simplistic utilization of generated data will constrain the model's ability to assimilate the identity in the invisible references. Therefore, we combine the generated videos with the static videos $V_{id}$ inflated from reference images provided by the user, forming an base video dataset repository $\mathcal{B} = V_t \cup V_s \cup V_{id}$.

\subsection{Customized Video Reward}
\label{custom}
According to the goal of our customization, we propose to evaluate the videos in the base dataset repository with three customized rewards: identity consistency, dynamic degree, and prompt following, with all scores ranging from 1 to 10. First, we use the pretrained ID Encoder~\cite{deng2019arcface} to assess the mean identity consistency $R_{id}$ of the videos with the reference images, which indicates the ability of the video to maintain the identity. Secondly, we use the RAFT model~\cite{teed2020raft} to analyze optical flow and calculate the motion intensity $R_{dy}$ between consecutive frames, which scores the dynamic degree of the videos, reflecting the level of dynamism in the video. Finally, we use VLM~\cite{bai2023qwenvlversatilevisionlanguagemodel} to evaluate prompt following reward $R_{sem}$, which further assess the original dynamics of the video in terms of the semantic alignment with the given prompts.

\subsection{Hybrid Pair Selection}
\label{hybrid}

To address the issues of identity degradation and dynamic reduction, we aim to construct the video pairs for preference optimization, which is essential to consider both identity and dynamics. To this end, we introduce a hybrid sampling strategy to construct identity-preferred and dynamic-preferred training pairs for two distinct stages, which is illustrated in~\cref{fig:method}. 
In the first stage, we exclusively utilize $V_{id}$ and $V_s$ from the video libraries, selecting video pairs $P_{id}$ with significant differences in identity consistency scores while permitting a certain degree of tolerance for dynamic reward. Clearly, the selected video pairs in this stage will guide the model toward maintaining a high level of identity consistency. 

In the second stage, we select dynamic-preferred video pairs $P_{dy}$ from $V_s$ and $V_t$ based on a Frontier-based sampling method. Specifically, we first find the upper and lower Pareto frontier set using a non-dominated sorting algorithm~\cite{deb2002fast}. For all videos from $V_s$ and $V_t$, we define $a$ dominates $b$ if and only if $R(a) > R(b)$ for $\forall R \in \{R_{id}, R_{dy}, R_{sem}\}$. Then we can find the non-dominated set whose elements are not be dominated by any other videos. Conversely, we can also identify the dominated set and then construct the preference pairs from them.

After that, we rank all video pairs from \( P_{dy} \) based on the differences in identity consistency scores and retain only the top 100 pairs as our preference data. By combining these video pairs with \( P_{id} \), we construct \( P = P_{dy} \cup P_{id} \), forming training pairs that prioritize both identity consistency and dynamic effects, effectively addressing the two primary limitations of personalized customization tasks.

\subsection{Hybrid Preference Optimization }

Based on the construction pipeline above, our pairwise preference video data can be represented as $P = \{(c, v_0^w, v_0^l)\}$
where each sample contains and a pair of videos \( (v_0^w, v_0^l) \) filtered by~\cref{hybrid} and their text prompts \( c \), with \( v_0^w \succ v_0^l \) indicating that \( v_0^w \) better aligns with human preferences. Now, our goal is to train a new model \( p_\theta \). To measure the quality of the complete generation path, we define a reward function \( R^*(c, v_{0:T}) \) and further derive the expected reward \( r(c, v_0) \) given \( c \) and \( v_0 \):
\begin{equation} \label{eq:expected_reward}
r(c, v_0) = \mathbb{E}_{p_\theta(v_{1:T} \mid v_0, c)} [R^*(c, v_{0:T})].
\end{equation}
Following previous work~\cite{liu2024videodpo,jiang2025huvidpo}, we substitute the upper bound of the KL divergence~\cite{kullback1951information} and the reward function \( r(c, v_0) \) into~\cref{max}, yielding:
\begin{equation} \label{eq:kl_maximization}
\begin{aligned}
\max_{p_\theta} &\mathbb{E}_{c \sim \mathcal{D}_c, v_{0:T} \sim p_\theta(v_{0:T} \mid c)} [r(c, v_0)] \\
&- \beta \mathbb{D}_{KL} \Big[p_\theta(v_{0:T} \mid c) \parallel p_{\text{ref}}(v_{0:T} \mid c)\Big].
\end{aligned}
\end{equation}

To optimize this objective, the conditional distribution \( p_\theta(x_{0:T}) \) is directly used. Similar to~\cref{www}, the final loss function can be defined. By applying Jensen's inequality~\cite{jensen1906fonctions},
the final upper-bound loss function \( \mathcal{L}_{\text{HPO}}(\theta) \) is expressed as:
\begin{equation}
\label{eq:3}
\begin{aligned}
\mathcal{L}_{\text{HPO}}(\theta) \leq &-\mathbb{E}_{(v_0^w, v_0^l) \sim D} \, \mathbb{E}_{v_{1:T}^w \sim p_\theta(v_{1:T} | v_0^w), \, v_{1:T}^l \sim p_\theta(v_{1:T} | v_0^l)} \\
& \ \ \ \ \ \ \ \ \ \ \ \ \ \ \ \ \ \ \ \ \ \log \sigma \left( \beta \left( \Delta(v_{0:T}^w) - \Delta(v_{0:T}^l) \right) \right),
\end{aligned}
\end{equation}
where \(\Delta(v_{0:T})\) is defined as:
\begin{equation}
\label{eq}
\Delta(v_{0:T}) = \log \frac{p_\theta(v_{0:T})}{p_{\text{ref}}(v_{0:T})}.
\end{equation}

Due to the complexity of calculating high-dimensional video sequence probabilities in video generation tasks, we introduce an approximate posterior distribution \( q(v_{1:T} \mid v_0) \) to estimate \( p_\theta(v_{0:T}) \). A more detailed derivation can be found in the supplementary material. Combining this with the evidence lower bound (ELBO) method~\cite{jordan1999introduction}, we can transform the probability distribution of video generation into a KL divergence representation:
\begin{equation} \label{eq:delta_kl}
\Delta(v_{0:T}) = -\mathbb{D}_{KL}^{\theta} + \mathbb{D}_{KL}^{\text{ref}} + C.
\end{equation}
Furthermore, by relating KL divergence to noise prediction, we can rewrite it as follows:
\begin{equation}
\label{eq:9}
\mathbb{D}_{KL}^{\theta} \propto ||\epsilon - \epsilon_\theta(v_t, t)||^2 
\quad \mathbb{D}_{KL}^{\text{ref}} \propto ||\epsilon - \epsilon_\text{ref}(v_t, t)||^2.
\end{equation}
Finally, the complete HPO loss function can be mathematically represented as follows:
\begin{equation}
\label{lab:2}
\begin{aligned}
\mathcal{L}_{\text{HPO}}(\theta) = & \, \mathbb{E}_{(v_{0}^w, v_{0}^l) \sim \mathcal{D}, \, t \sim \{1..T\}} \Bigg[\beta \log \sigma \Bigg( \\
& \quad \left( ||\epsilon_w - \epsilon_\theta(v^w_t, t)||^2 - ||\epsilon_w - \epsilon_{\text{ref}}(v^w_t, t)||^2 \right) \\
& \quad - \left( ||\epsilon_l - \epsilon_\theta(v^l_t, t)||^2 - ||\epsilon_l - \epsilon_{\text{ref}}(v^l_t, t)||^2 \right) \Bigg) \Bigg].
\end{aligned}
\end{equation}

\begin{figure*}[t]
\centering
\includegraphics[width=\linewidth]{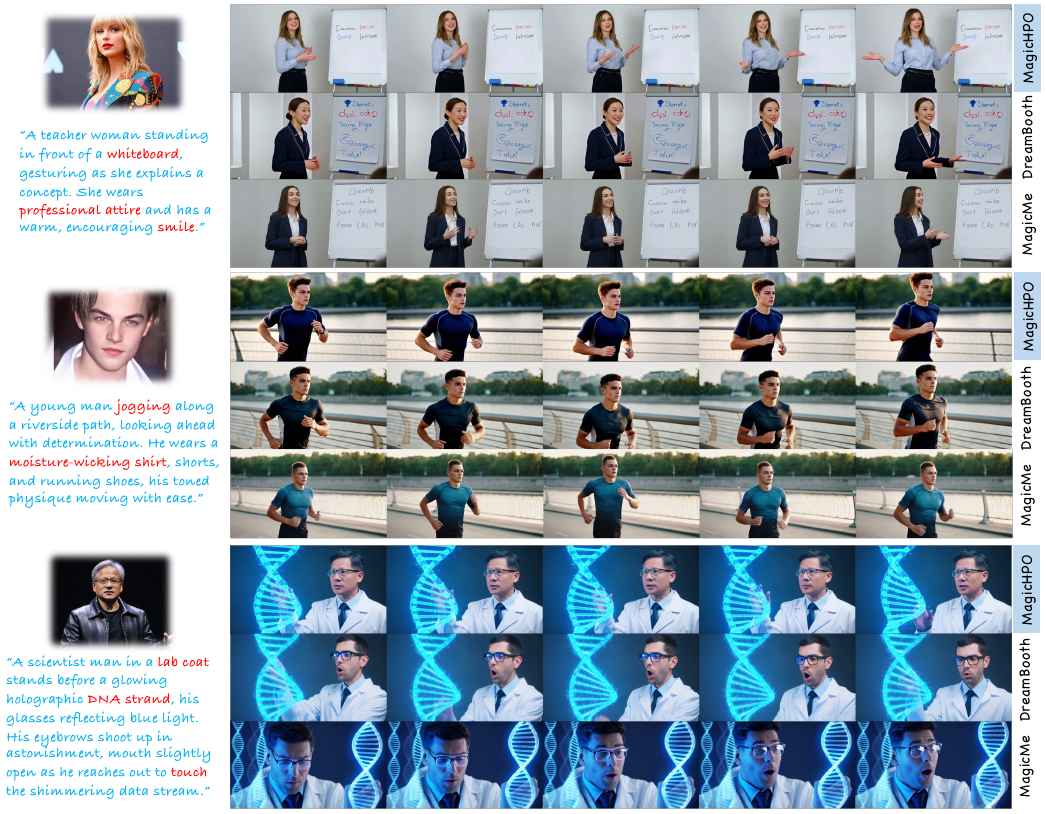}
\vspace{-0.6cm}
\caption{\textbf{Qualitative comparison with tuning-based methods.} As observed, both Dreambooth and MagicMe suffer from inferior ID fidelity, while our method maintains consistent identity and natural dynamics.}
\label{fig:result1}
\vspace{-0.3cm}
\end{figure*}

\begin{figure*}[t]
\centering
\includegraphics[width=\linewidth]{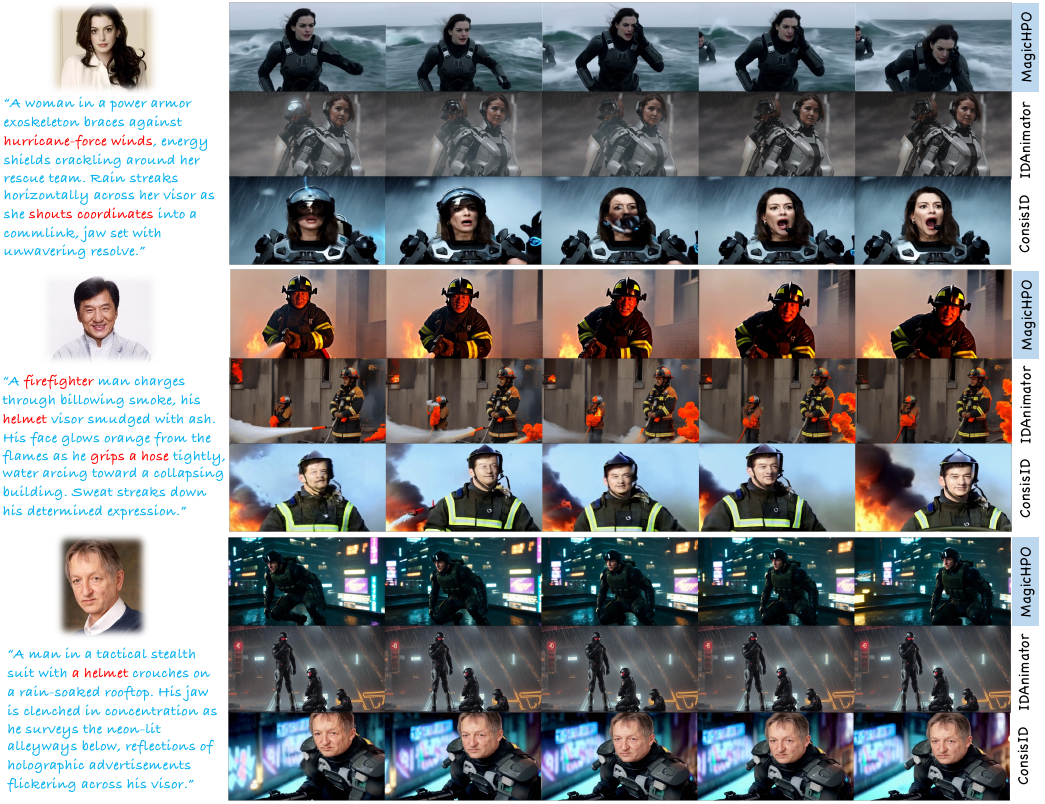}
\vspace{-0.4cm}
\caption{\textbf{Qualitative comparison with tuning-based methods.} As shown, ID-Animator suffers from poor identity consistency and video quality. While ConsisID improves identity fidelity to some extent, it exhibits severe \textit{copy-paste} artifacts, demonstrating unnatural motion dynamics and text alignment, as seen in the last example with the \textit{helmet}. In contrast, our method achieves strong performance in identity consistency, motion dynamics, and text alignment, significantly outperforming the baseline approaches.}
\label{fig:result2}
\vspace{-0.2cm}
\end{figure*}

\begin{figure}[t]
\centering
\includegraphics[width=\linewidth]{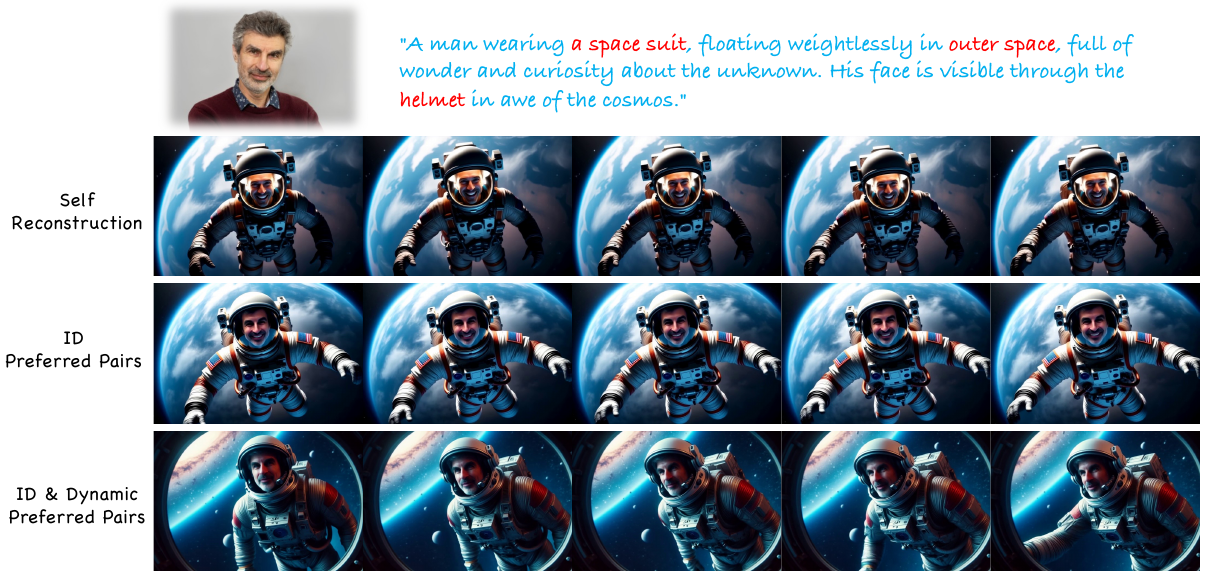}
\vspace{-0.3cm}
\caption{\textbf{Ablation study for the hybrid pair selection.} Compared to the self-reconstruction approach, training with identity-preferred pairs significantly enhances the identity consistency. On the other hand, the incorporation of dynamic-preferred pairs markedly improves the dynamics of the generated outcomes.}
\label{fig:pair}
\vspace{-0.2cm}
\end{figure}

\begin{figure}[t]
\centering
\includegraphics[width=\linewidth]{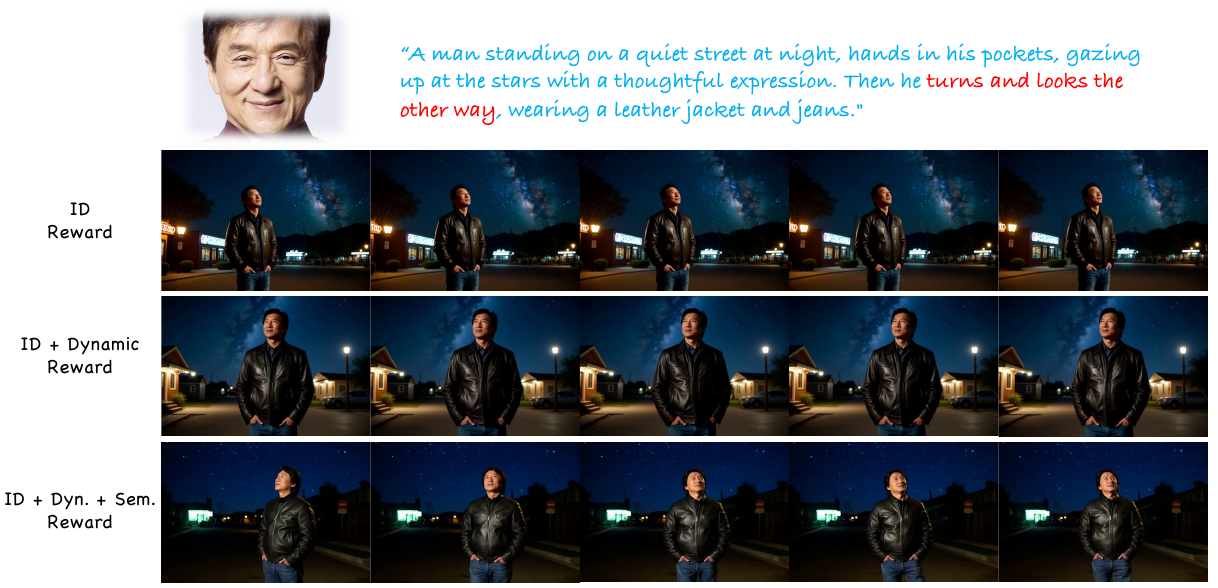}
\vspace{-0.3cm}
\caption{\textbf{Ablation study for the customized video rewards.} While the ID reward encourages the model to learn consistent identity features, the addition of the dynamic reward leads to generated results with significantly improved motion dynamics. Furthermore, incorporating a semantic reward can effectively enhance video dynamics to some extent and improve prompt-following capabilities, such as the prompts involving \textit{turning} motions.}
\label{fig:reward}
\vspace{-0.2cm}
\end{figure}

\section{Experiments}
\subsection{Experimental Setup}
\noindent
\textbf{Implementation details.}
We utilize the recently developed Text-to-Video DiT model, HunyuanVideo~\cite{kong2024hunyuanvideo}, as our foundational model. For the training process, we employ the AdamW optimizer configured with a learning rate of 2e-5 and a weight decay parameter of 1e-4. We first fine-tune the model for 1000 steps in the initial training stage. Then we use our training method to optimize for 5000 steps, which maintains parity with the total number of training steps used in the baseline method, ensuring a fair comparison.
Please refer to Appendix for more details.

\noindent
\textbf{Baselines.}
Our comparative analysis benchmarks the proposed method against MagicMe~\citep{ma2024magic}, a recent identity-specific T2V customization method, as well as DreamBooth with LoRA~\citep{hu2021lora}. For a fair comparison, we apply both DreamBooth and MagicMe on the HunyuanVideo model. Additionally, we compare the results with encoder-based methods like IDAnimator~\citep{he2024id} and ConsisID~\cite{yuan2024identity}, both trained on meticulously collected large-scale video datasets.

\noindent
\textbf{Evaluation.}
Our evaluation dataset consists of 40 characters ensuring demographic representation, with 40 action-specific prompts for comprehensive motion evaluation. 
The evaluation framework employs the standardized VBench benchmark~\cite{huang2024vbench} for quantitative assessment of dynamic degree and text alignment. To measure identity preservation, we implement facial recognition embedding similarity metrics~\cite{an2021partial} complemented by specialized facial motion analysis protocols. Additionally, we calculate other well-established metrics for comprehensive video evaluation including Temporal Consistency, Image CLIP Score, and FVD. Unless otherwise specified, we generate 61-frame videos during the inference phase.

\subsection{Main Results}
\noindent
\textbf{Qualitative results.}
We present a qualitative assessment comparing \modulespace with baseline methods. As illustrated in \cref{fig:result1}, both DreamBooth and MagicMe exhibit suboptimal identity fidelity, primarily due to their self-reconstruction training approach on reference images, which creates a significant discrepancy between the training 1-frame image and the multi-frame videos generated during inference. Conversely, our \modulespace achieves superior ID fidelity while preserving the integrity of motion dynamics.
To further validate the robustness of our methodology, we also train using a single reference image for a fair comparison with encoder-based methods. As shown in \cref{fig:result2}, ID-Animator struggles with maintaining identity consistency and delivering high-quality video outputs. Although ConsisID enhances identity fidelity to a certain degree, it introduces noticeable \textit{copy-paste} effects, resulting in unnatural motion dynamics and suboptimal text alignment, as illustrated in the final example featuring the \textit{helmet}. In contrast, our results further highlight the advantages of \module, showcasing its robustness in achieving high ID fidelity while maintaining promising motion dynamics.

\begin{figure}[t]
\centering
\includegraphics[width=\linewidth]{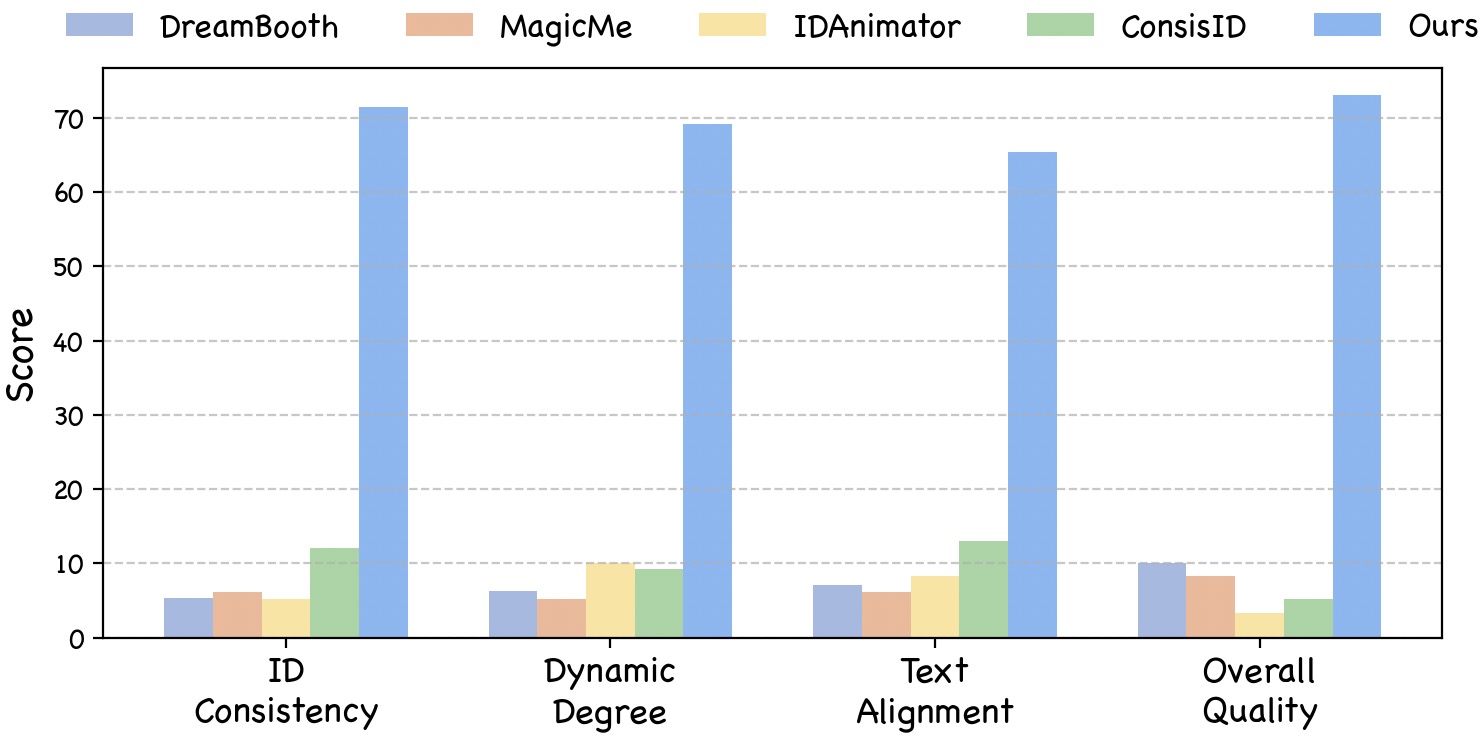}
\vspace{-0.3cm}
\caption{\textbf{User Study.} }
\label{fig:user}
\end{figure}

\noindent
\textbf{Quantitative results.}
We present the quantitative results in \cref{tab:result}. As observed, DreamBooth exhibits inferior face similarity due to its self-reconstruction training approach. MagicMe achieves better face similarity but at the cost of degraded motion dynamics and prompt adherence.
Leveraging extensive video datasets, ID-Animator and ConsisID attain superior facial similarity and motion dynamics, albeit with diminished prompt adherence and temporal consistency.
In comparison, our \modulespace method outperforms prior approaches, particularly in facial similarity and dynamic quality. It achieves exceptional identity fidelity while effectively preserving the original text-to-video model's capabilities, aligning with the qualitative observations.

\begin{table*}[ht]
    \tablestyle{3pt}{1.15}
    \centering
    \begin{tabular}{ccccccc} 
        \toprule
\textbf{Method} & \textbf{Face Sim.} ($\uparrow$) & \textbf{Dyna. Deg.} ($\uparrow$) & \textbf{T. Cons.} ($\uparrow$) & \textbf{CLIP-T} ($\uparrow$)& \textbf{CLIP-I} ($\uparrow$) & \textbf{FVD} ($\downarrow$)\\
\midrule
DreamBooth \cite{ruiz2023dreambooth} &0.276 &5.690 &0.9922   &25.83 &46.55 &1423.55\\
MagicMe \cite{ma2024magic} &0.322 &5.332 &0.9924  &25.42 &62.45  &1438.66\\
IDAnimator \cite{he2024id} &0.433 &10.33 &0.9938   &25.21 &49.33 &1558.33 \\
ConsisID \cite{yuan2024identity} &0.482 &9.26  &0.9811  &26.12 &63.88 &1633.21\\
\textbf{\module} &\textbf{0.600} &\textbf{14.42} &  \textbf{0.9933} &\textbf{26.28} & \textbf{78.83} &\textbf{1228.33} \\
        \bottomrule
    \end{tabular}
\vspace{-0.2cm}
\caption{\textbf{Quantitative comparison.} We conduct a comprehensive comparison including the ability to achieve high ID fidelity (\ie, Face Similarity and CLIP-I), dynamic degree, temporal consistency, text alignment (\ie, CLIP-T), and distribution distance (\ie, FVD). 
}
\label{tab:result}
\vspace{-0.2cm}
\end{table*}

\noindent
\textbf{User study.}
To further evaluate the effectiveness of our methodology, we conduct a human-centric assessment, comparing our approach with existing text-to-video identity customization techniques. We recruit 25 evaluators to assess 40 sets of generated video results. For each set, we present reference images alongside videos produced using identical seeds and textual prompts across various methods. The quality of the generated videos is evaluated based on four criteria: ID Consistency (the degree of resemblance between the generated subject and the reference image), Dynamic Degree (the extent of motion dynamics within the video), Text Alignment (the fidelity of the video to the textual prompt), and Overall Quality (the general user satisfaction with the video quality). As depicted in \cref{fig:user}, our \modulespace achieves higher user preference across all evaluative dimensions, underscoring its superior effectiveness.

\subsection{Ablation Study}
\noindent
\textbf{Effects of hybrid pair selection.}
To validate the impact of the proposed strategy for preference pair selection, we conduct a detailed ablation study in \cref{fig:pair} and \cref{tab:pair}. In contrast to the self-reconstruction method, utilizing identity-preferred pairs during preference optimization significantly enhances identity consistency in the results. Additionally, integrating dynamic-preferred pairs notably improves the motion dynamics of the generated outputs, demonstrating the effectiveness of our hybrid pair selection.

\noindent
\textbf{Effects of customized video reward.}
To verify the effects of our customized video reward, we conduct experiments to compare the results of different rewards. As shown in \cref{fig:reward} and \cref{tab:reward}, while the ID reward encourages the model to acquire more consistent identity characteristics, it somewhat neglects the dynamic quality of the video. In comparison, the inclusion of the dynamic reward results in generated outcomes with enhanced motion dynamics. Additionally, integrating a semantic reward can augment video dynamics and also improves prompt-following abilities.

\begin{table}[t]
    \tablestyle{3pt}{1.15}
    \centering
    \begin{tabular}{cc|ccc}
    \toprule
        ID pairs & Dynamic pairs & \textbf{Face} ($\uparrow$)  & \textbf{Dynamic} ($\uparrow$)& \textbf{CLIP-T}  ($\uparrow$)\\
            \midrule
        & &0.276 &5.690  &25.83\\
        \checkmark& &\textbf{0.605} &7.382 &25.94\\
        \checkmark&\checkmark &0.600 &\textbf{14.42} &\textbf{26.28}\\
    \bottomrule
    \end{tabular}
    \caption{Quantitative ablation study of hybrid pair selection.}
\label{tab:pair}
\vspace{-0.2cm}
\end{table}

\begin{table}[t]
    \tablestyle{3pt}{1.15}
    \centering
    \begin{tabular}{ccc|ccc}
    \toprule
        ID & Dynamic & Semantic & \textbf{Face} ($\uparrow$) & \textbf{Dynamic} ($\uparrow$) & \textbf{CLIP-T}  ($\uparrow$)\\
            \midrule
        \checkmark&& &0.598 &6.332  &24.92\\
    \checkmark&\checkmark& &\textbf{0.607} &12.33  &25.73\\
        \checkmark&\checkmark&\checkmark &0.600 &\textbf{14.42} &\textbf{26.28}\\
    \bottomrule
    \end{tabular}
    \caption{Quantitative ablation study of customized video reward.}
\label{tab:reward}
\vspace{-0.2cm}
\end{table}
\section{Conclusion \& Limitation}
In this paper, we present \module, a novel framework designed to address the significant challenges in identity-specific video generation. By introducing a hybrid customized preference optimization, our method effectively maintains consistent identity fidelity and preserves natural motion dynamics, overcoming the limitations of traditional self-reconstruction techniques. Through a two-stage hybrid sampling strategy, we construct identity-preferred and dynamic-preferred training pairs, ensuring robust identity learning and enhanced video dynamics. Our framework demonstrates superior performance in generating high-quality personalized videos, which holds substantial promise for applications in the film and television industries, paving the way for more realistic and engaging content creation.
Future work will explore further refinements and broader applications of our approach in diverse multimedia domains.

However, our method also has some limitations. For example, it focuses on single-person consistent identity video generation, but fails to generate customized videos that contain multiple identities. One possible solution is to introduce a reward mechanism tailored for multi-person video generation, which will be explored in our future work.

\clearpage
{
    \small
    \bibliographystyle{ieeenat_fullname}
    \bibliography{main}
}


\setcounter{page}{1}
\appendix
\onecolumn 
\begin{center}
\Large
\textbf{MagicID: Hybrid Preference Optimization for ID-Consistent and
Dynamic-Preserved Video Customization}\\
\vspace{0.5em}Supplementary Material \\
\end{center}

\section{Implement Details}
To construct the preference video repository, we generate 20 prompts using LLM model. Then we sample 100 and 20 videos with fine-tuned T2V model and the initial T2V model respectively, which are incorporated with the static videos inflated from the reference images. After the pair selection, we sample the top 100 pairs as our preference data.
In the training stage, we employ the AdamW optimizer configured with a learning rate of 2e-5 and a weight decay parameter of 1e-4. We first fine-tune the model for 1000 steps in the initial training stage. Then we use our training method to optimize for 5000 steps.
During inference, we use 50 steps of DDIM sampler and classifier-free guidance with a scale of 7.5. We generate 61-frame videos with 720 × 1280 spatial resolution. 
All experiments are conducted on a single NVIDIA H00 GPU. 

\section{Baseline Details}
We compare our method with both optimization methods, such as Magic-Me and Dreambooth with LoRA, and encoder-based methods such as IDAnimator and ConsisID. Specifically, Magic-Me is a recent T2V customization method that trains extended keywords and injects it into HunyuanVideo. Besides, we compare with Dreambooth-LoRA, which uses traditional reconstructive loss during training. For a fair comparison, we train them for the same total steps with our method. 

\section{DPO Objective Function}
This section provides a systematic derivation of the Direct Preference Optimization (DPO) formula, detailing the derivation of the DPO objective function and constructing a preference-driven optimization framework based on the reward function and KL divergence. The goal of DPO is to maximize rewards while aligning the policy with a baseline model. The objective function is defined as:
\begin{equation}
\max_\pi \mathbb{E}_{x \in X, y \in \pi} \left[ r(x, y) \right] - \beta \cdot \mathbb{D}_{\text{KL}} \left[ \pi(y | x) || \pi_{\text{ref}}(y | x) \right],
\end{equation}
where \( D_{\text{KL}} \) denotes the Kullback-Leibler divergence between the learned policy \( \pi \) and a reference policy \( \pi_{\text{ref}} \), enforcing consistency with the baseline model. To simplify, this objective is reformulated as a minimization problem:
\begin{equation}
\min_\pi \mathbb{E}_{x \in X, y \in \pi} \left[ \log \frac{\pi(y | x)}{\pi^*(y | x)} - \log Z(x) \right],
\end{equation}
where \( Z(x) \) is defined as:
\begin{equation}
Z(x) = \sum_y \pi_{\text{ref}}(y | x) \exp \left( \frac{1}{\beta} r(x, y) \right).
\end{equation}
This reformulation leads to the final optimization objective:
\begin{equation}
\min_\pi \mathbb{E}_{x \sim D} \left[ \mathbb{D}_{\text{KL}}(\pi(y | x) || \pi^*(y | x)) \right].
\end{equation}
Under the minimization of KL divergence, the policy \( \pi(y | x) \) adheres to the following form:
\begin{equation}
\pi(y | x) = \pi^*(y | x) = \frac{1}{Z(x)} \pi_{\text{ref}}(y | x) \cdot \exp \left( \frac{1}{\beta} r(x, y) \right).
\end{equation}
Reversing this equation yields the reward function:
\begin{equation}
r^*(x, y) = \beta \log \frac{\pi(y \mid x)}{\pi_{\text{ref}}(y \mid x)}.
\end{equation}
Incorporating the Bradley-Terry model, the cross-entropy loss function \(\mathcal{L}\) is defined, which quantifies the difference between the preferred and non-preferred responses. This loss function is essential for deriving the gradient necessary to optimize the DPO objective:
\begin{equation}
\mathcal{L} = -\mathbb{E}_{(x, y_w, y_l) \sim D} \left[ \ln \sigma \left( \beta \log \frac{\pi(y_w | x)}{\pi_{\text{ref}}(y_w | x)} - \beta \log \frac{\pi(y_l | x)}{\pi_{\text{ref}}(y_l | x)} \right) \right],
\end{equation}
where \( \sigma \) denotes the sigmoid function, which maps the difference in log-probabilities to a range of \([0, 1]\). Differentiating \(\mathcal{L}\) provides the gradient needed to optimize the DPO objective with respect to the preference data.

\section{Applying DPO Strategy into our MagicID}

In adapting Dynamic Preference Optimization (DPO) to our MagicID task, we consider a pairwise preference video data \( P = \{(c, v_0^w, v_0^l)\} \). In this dataset, each example contains their text prompts \( c \) and a pair of videos \( (v_0^w, v_0^l) \) generated by a reference model \( p_{\text{ref}} \), where \( v_0^w \succ v_0^l \) indicates that humans prefer \( v_0^w \) over \( v_0^l \). The goal of DPO is to train a new model \( p_\theta \) so that its generated videos align with human preferences rather than merely imitating the reference model. However, directly computing the distribution \( p_\theta(v_0 | c) \) is highly complex, as it requires marginalizing over all possible generation paths \( (v_1, \ldots, v_T) \) to produce \( v_0 \), which is practically infeasible.

To address this challenge, researchers leverage Evidence Lower Bound (ELBO) by introducing latent variables \( v_{1:T} \). The reward function \( R(c, v_{0:T}) \) is defined to measure the quality of the entire generation path, allowing the expected reward \( r(c, v_0) \) for given \( c \) and \( v_0 \) to be formulated as:

\begin{equation}
r(c, v_0) = \mathbb{E}_{p_\theta(v_{1:T} | v_0, c)} [R(c, v_{0:T})]
\end{equation}

In DPO, a KL regularization term is also included to constrain the generated distribution relative to the reference distribution. Here, an upper bound on the KL divergence is used, converting it to a joint KL divergence:

\begin{equation}
\mathbb{D}_{KL}[p_\theta(v_{0:T} | c) \parallel p_{\text{ref}}(v_{0:T} | c)]
\end{equation}

This upper bound ensures that the distribution of the generated model \( p_\theta(v_{0:T} | c) \) remains consistent with the reference model \( p_{\text{ref}}(v_{0:T} | c) \), preserving the model’s generation capabilities while optimizing human preference alignment. Plugging in this KL divergence upper bound and the reward function \( r(c, v_0) \) into the objective function, we obtain:

\begin{equation}
\max_{p_\theta} \mathbb{E}_{c \sim \mathcal{D}_c, v_{0:T} \sim p_\theta(v_{0:T} | c)} [r(c, v_0)] - \beta \mathbb{D}_{KL} [p_\theta(v_{0:T} | c) \parallel p_{\text{ref}}(v_{0:T} | c)]
\end{equation}

The definition of this objective function is optimized over the path \( v_{0:T} \). Its primary goal is to maximize the reward for the reverse process \( p_\theta(v_{0:T}) \) while maintaining distributional consistency with the original reference reverse process. To optimize this objective, the conditional distribution \( p_\theta(v_{0:T}) \) is directly used. The final DPO-MagicID loss function \( L_{\text{HPO}}(\theta) \) is expressed as follows:

\begin{equation}
\mathcal{L}_{\text{HPO}}(\theta) = -\mathbb{E}_{(v_0^w, v_0^l) \sim P} \log \sigma \left( \beta \mathbb{E}_{v_{1:T}^w \sim p_\theta(v_{1:T} | v_0^w), \, v_{1:T}^l \sim p_\theta(v_{1:T} | v_0^l)} \left[ \log \frac{p_\theta(v_{0:T}^w)}{p_{\text{ref}}(v_{0:T}^w)} - \log \frac{p_\theta(v_{0:T}^l)}{p_{\text{ref}}(v_{0:T}^l)} \right] \right)
\end{equation}

By applying Jensen's inequality, the expectation can be moved outside of the \( \log \sigma \) function, resulting in an upper bound. This simplifies the formula and facilitates optimization. After applying Jensen's inequality, the upper bound of the loss function is given by:

\begin{equation}
\label{wewrr}
\mathcal{L}_{\text{HPO}}(\theta) \leq -\mathbb{E}_{(v_0^w, v_0^l) \sim P} \, \mathbb{E}_{v_{1:T}^w \sim p_\theta(v_{1:T} | v_0^w), \, v_{1:T}^l \sim p_\theta(v_{1:T} | v_0^l)} \, \log \sigma \left( \beta \left[ \log \frac{p_\theta(v_{0:T}^w)}{p_{\text{ref}}(v_{0:T}^w)} - \log \frac{p_\theta(v_{0:T}^l)}{p_{\text{ref}}(v_{0:T}^l)} \right] \right)
\end{equation}


To handle the complexity of calculating high-dimensional video sequence probabilities with a total of \( T = 1000 \) time steps, we employ an approximation approach. We introduce an approximate posterior \( q(v_{1:T} | v_0) \) for the subsequent time steps and utilize the Evidence Lower Bound (ELBO) to approximate $\log p_\theta(v_{0:T})$. Then, by expressing \( p_\theta(v_{0:T}) \) and \( q(v_{1:T} | v_0) \) as products of conditional probabilities at each time step, we achieve a stepwise sampling approach. The final approximate expression is:
\begin{equation}
\label{KL1}
\begin{split}
\log p_\theta(v_{0:T}) \approx \mathbb{E}_{q(v_t | v_{t-1}), t \sim \{1..T\}} 
\Bigg[ \log \frac{p_\theta(v_0)}{q(v_0)} \\
+ \log \frac{p_\theta(v_t | v_{t-1})}{q(v_t | v_{t-1})} \Bigg].
\end{split}
\end{equation}
Since \( q(v_t | v_{t-1}) \) is a conditional probability distribution that generally sums to 1, the KL divergence can be expressed as:
\begin{equation}
\label{KL2}
\mathbb{D}_{KL}\left( q(v_t | v_{t-1}) \, \| \, p_\theta(v_t | v_{t-1}) \right) =  \log \frac{q(v_t | v_{t-1})}{p_\theta(v_t | v_{t-1})}.
\end{equation}
Based on~\cref{KL1,KL2}, we rewrite \( \log p_\theta(v_{0:T}) \) as:
\begin{equation}
\label{eq:6}
\begin{split}
\log p_\theta(v_{0:T}) \approx \mathbb{E}_{q(v_{1:T} | v_0), t \sim \{1..T\}} \left[ \log \frac{p_\theta(v_0)}{q(v_0)} \right] \\ 
- \mathbb{D}_{KL}\left( q(v_t | v_{t-1}) \, \| \, p_\theta(v_t | v_{t-1}) \right).
\end{split}
\end{equation}
Moreover, the derivation of \( \log p_{\text{ref}}(v_{0:T}) \) is consistent with that of \( \log p_\theta(v_{0:T}) \). Based o~\cref{eq:6}, we can rewrite \( \Delta(v_{0:T}) \) as:
\begin{equation}
\label{eq:7}
\log \frac{p_\theta(v_{0:T})}{p_{\text{ref}}(v_{0:T})} = -\mathbb{D}_{KL}^{\theta} + \mathbb{D}_{KL}^{\text{ref}} + C.
\end{equation}
By rewriting the KL divergence in terms of noise prediction, we can express it as follows:
\begin{equation}
\mathbb{D}_{KL}^{\theta} \propto ||\epsilon - \epsilon_\theta(v_t, t)||^2 
\quad \mathbb{D}_{KL}^{\text{ref}} \propto ||\epsilon - \epsilon_\text{ref}(v_t, t)||^2.
\end{equation}
Finally, based on~\cref{wewrr,eq:7,eq:9}, the complete form of the DPO loss function for our MagicID is:

\begin{equation}
\begin{aligned}
\mathcal{L}_{\text{HPO}}(\theta) = & \, \mathbb{E}_{(v_{0}^w, v_{0}^l) \sim \mathcal{D}, \, t \sim \{1..T\}} \Bigg[\beta \log \sigma \Bigg( \\
& \quad \left( ||\epsilon_w - \epsilon_\theta(v^w_t, t)||^2 - ||\epsilon_w - \epsilon_{\text{ref}}(v^w_t, t)||^2 \right) \\
& \quad - \left( ||\epsilon_l - \epsilon_\theta(v^l_t, t)||^2 - ||\epsilon_l - \epsilon_{\text{ref}}(v^l_t, t)||^2 \right) \Bigg) \Bigg].
\end{aligned}
\end{equation}

\section{More Results}
\begin{figure*}[t]
\centering
\includegraphics[width=.9\linewidth]{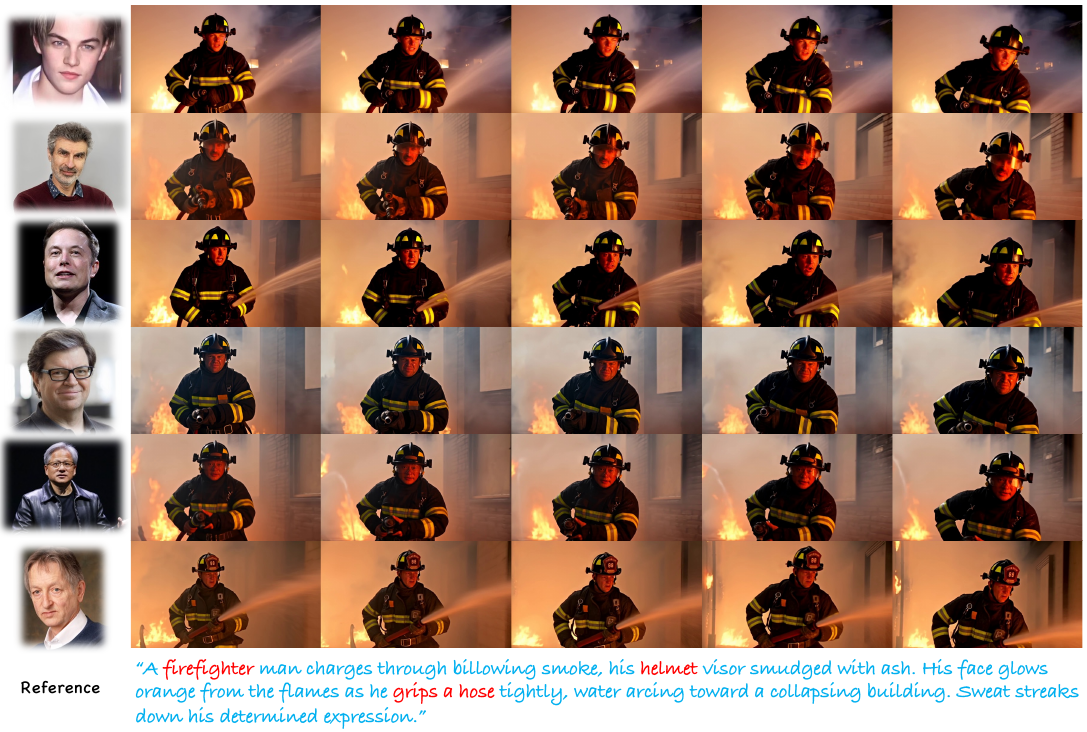}
\vspace{-0.2cm}
\caption{More results of \module.}
\label{fig:a5}
\vspace{-0.2cm}
\end{figure*}

\begin{figure*}[t]
\centering
\includegraphics[width=.9\linewidth]{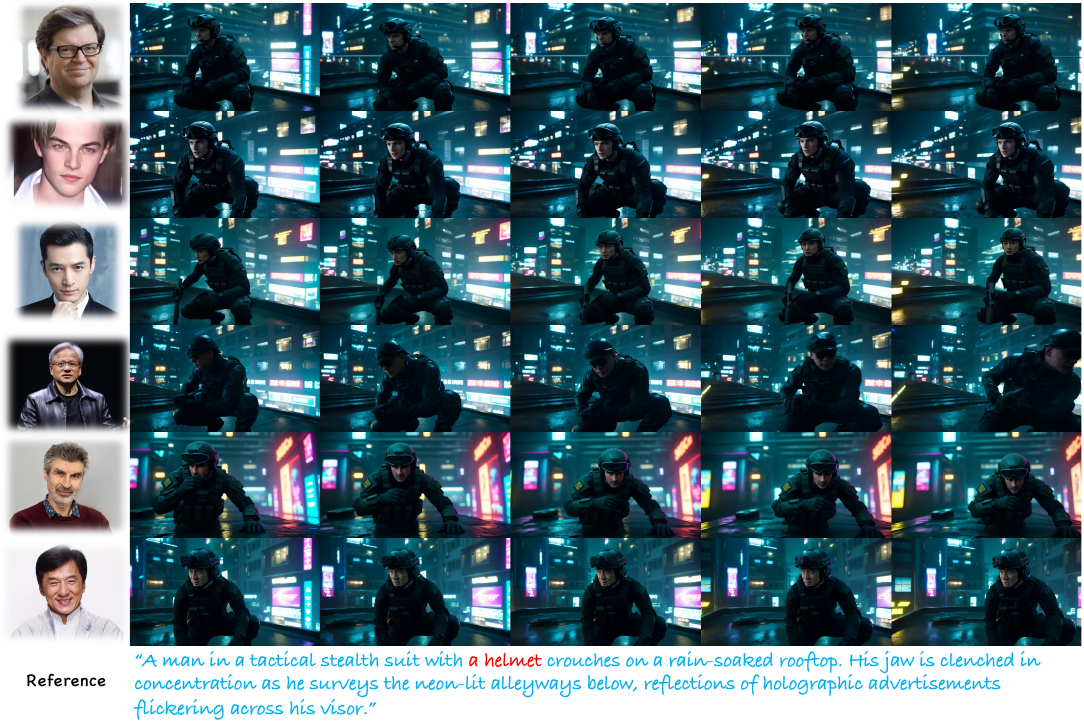}
\vspace{-0.2cm}
\caption{More results of \module.}
\label{fig:a6}
\vspace{-0.2cm}
\end{figure*}

\begin{figure*}[t]
\centering
\includegraphics[width=.9\linewidth]{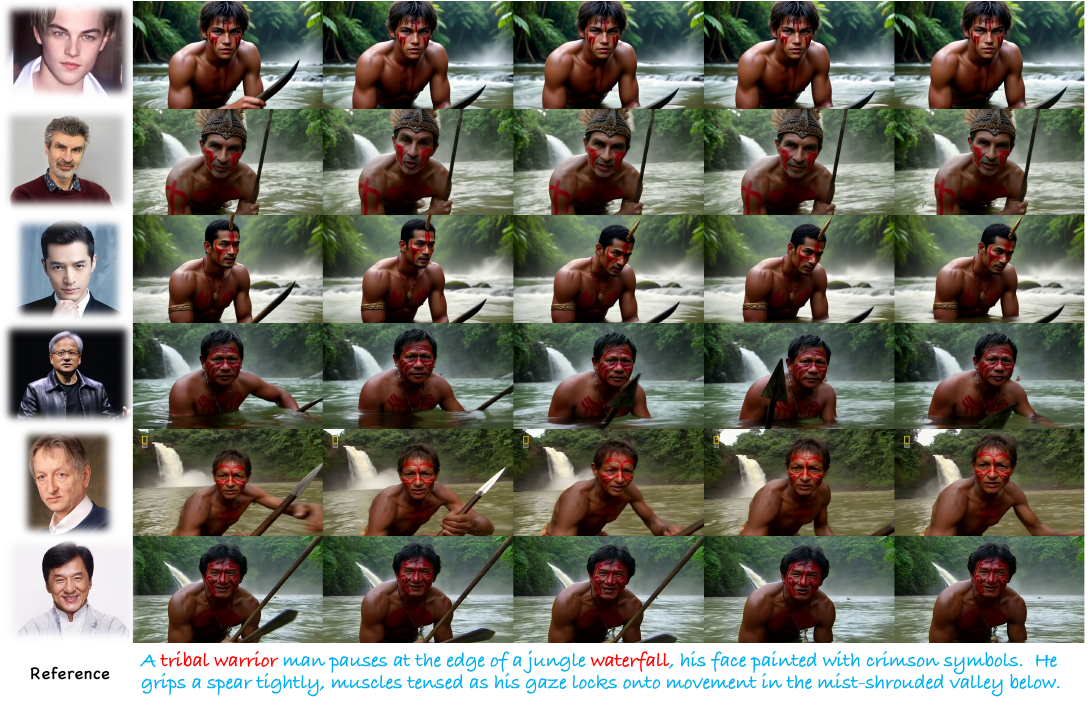}
\caption{More results of \module.}
\label{fig:a7}
\end{figure*}

\begin{figure*}[t]
\centering
\includegraphics[width=.9\linewidth]{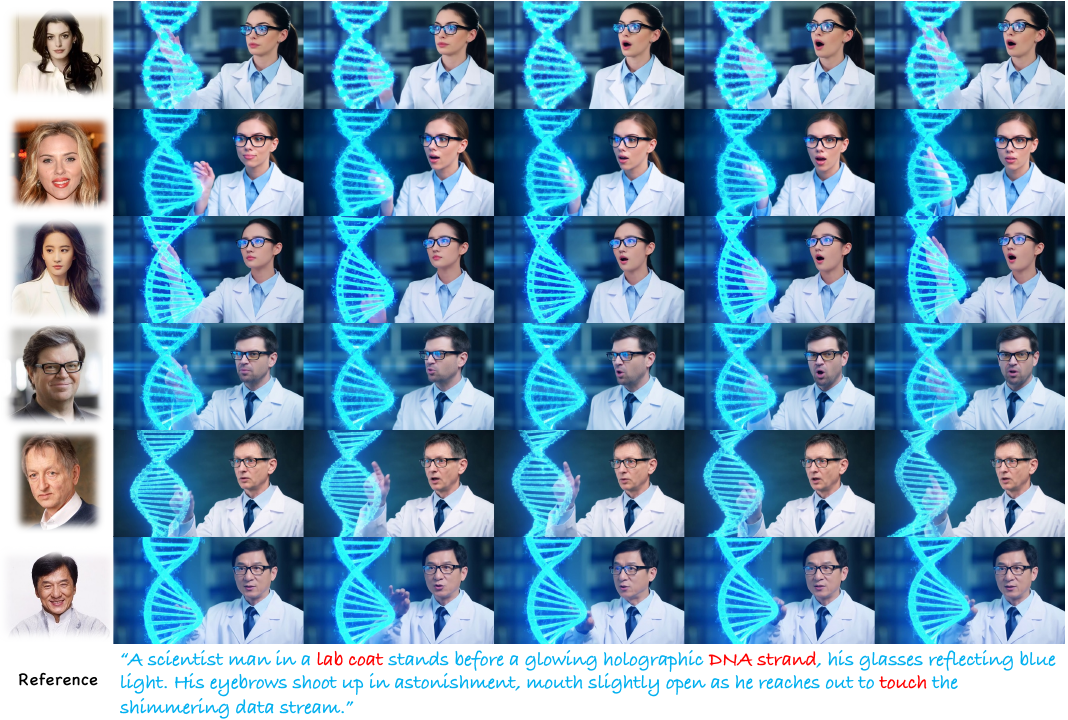}
\caption{More results of \module.}
\label{fig:a8}
\end{figure*}

\begin{figure*}[t]
\centering
\includegraphics[width=.9\linewidth]{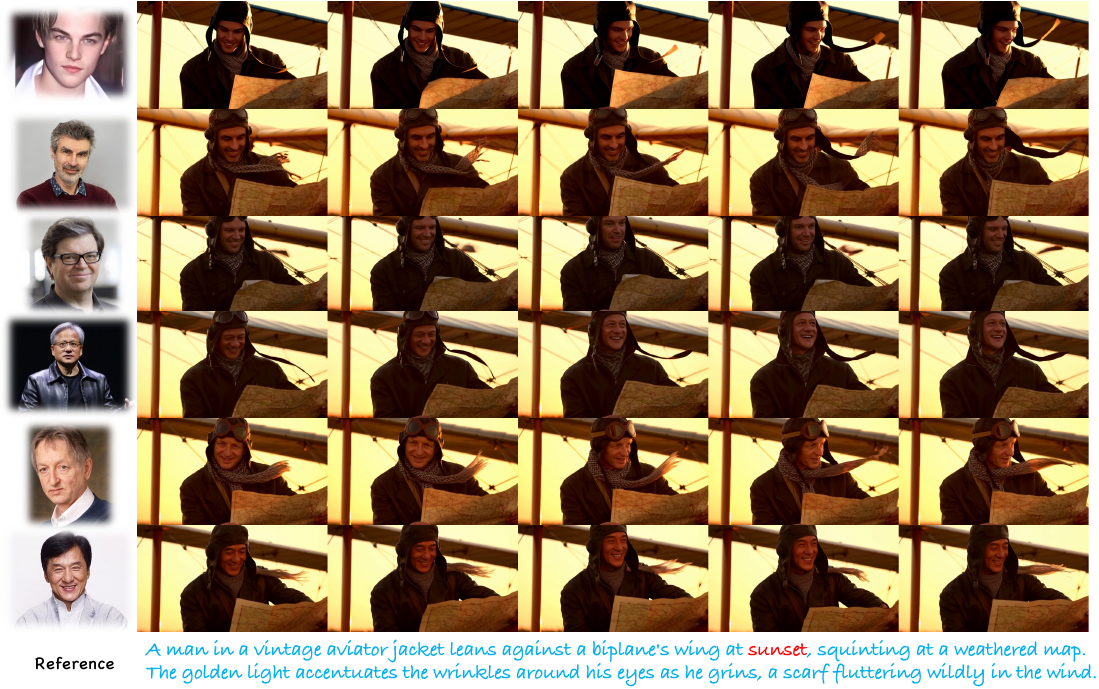}
\caption{More results of \module.}
\label{fig:a9}
\end{figure*}

As shown in \cref{fig:a5}, \cref{fig:a6}, \cref{fig:a7}, \cref{fig:a8}, and \cref{fig:a9}, we present more customization results of \module. They showcase it achieves consistent identity and preserves natural motion dynamics, which provides further evidence of its promising performance.

\section{Reproducibility Statement}
We make the following efforts to ensure the reproducibility of MagicID: (1) Our training and inference codes together with the trained model weights will be publicly available. (2) We provide training details in the appendix, which is easy to follow. (3) We provide the details of the human evaluation setups.
 
\section{Impact Statement}
Our main objective in this work is to empower novice users to generate visual content creatively and flexibly. However, we acknowledge the potential for misuse in creating fake or harmful content with our method. Thus, we believe it's essential to develop and implement tools to detect biases and malicious use cases to promote safe and equitable usage.
\end{document}